\documentclass[conference]{IEEEtran}

\IEEEoverridecommandlockouts
\usepackage{cite}
\usepackage{amsmath,amssymb,amsfonts}
\usepackage{algorithm}
\usepackage{graphicx}
\usepackage{textcomp}
\usepackage{xcolor}
\usepackage{float}
\usepackage[noend]{algpseudocode}
\usepackage{subfigure}

\usepackage{algorithmicx,algorithm}
\def\BibTeX{{\rm B\kern-.05em{\sc i\kern-.025em b}\kern-.08em
    T\kern-.1667em\lower.7ex\hbox{E}\kern-.125emX}}
\begin{document}

\title{Enhancing Text-to-Image Generation via End-Edge Collaborative Hybrid Super-Resolution}

\author{\IEEEauthorblockN{Chongbin~Yi, Yuxin Liang, Ziqi Zhou, and~Peng~Yang
\IEEEauthorblockA{School of Electronic Information and Communications, Huazhong University of Science and Technology, Wuhan, China\\
Email:\{chongbin\_y, yuxinliang, ziqi\_zhou, yangpeng\}@hust.edu.cn }}}


\maketitle

\begin{abstract}
Artificial Intelligence-Generated Content (AIGC) has made significant strides, with high-resolution text-to-image (T2I) generation becoming increasingly critical for improving users’ Quality of Experience (QoE). Although resource-constrained edge computing adequately supports fast low-resolution T2I generations, achieving high-resolution output still faces the challenge of ensuring image fidelity at the cost of latency. To address this, we first investigate the performance of super-resolution (SR) methods for image enhancement, confirming a fundamental trade-off that lightweight learning-based SR struggles to recover fine details, while diffusion-based SR achieves higher fidelity at a substantial computational cost. Motivated by these observations, we propose an end-edge collaborative generation-enhancement framework. Upon receiving a T2I generation task, the system first generates a low-resolution image based on adaptively selected denoising steps and super-resolution scales at the edge side, which is then partitioned into patches and processed by a region-aware hybrid SR policy. This policy applies a diffusion-based SR model to foreground patches for detail recovery and a lightweight learning-based SR model to background patches for efficient upscaling, ultimately stitching the enhanced ones into the high-resolution image. Experiments show that our system reduces service latency by 33\% compared with baselines while maintaining competitive image quality. 
\end{abstract}



\section{Introduction}
Recent years have witnessed the rapid development of Artificial Intelligence-Generated Content (AIGC), which has revolutionized content automation across diverse domains\cite{cao2025survey}. Within this broad landscape, Text-to-Image (T2I) generation stands out as a prominent paradigm, enabling users to guide image generation through simple textual prompts\cite{kong2025distributed}. In this context, with the growing demand for high-fidelity outputs, image resolution has emerged as a critical performance indicator\cite{wang2020deep}, significantly influencing the users' Quality of Experience (QoE) by determining perceptual quality and textural detail.

The prevailing architecture for AIGC services is cloud-based, where immense server resources support large-scale models such as Stable Diffusion 3 (8.1B) for high-resolution T2I generation \cite{gao2025characterizing}. This paradigm, however, suffers from significant communication latency due to long-distance data transmission. To mitigate this latency, edge-based AIGC has emerged as a promising alternative by deploying models closer to users \cite{zhuang2025qos,Liang2026TCCN}. But the resource constraints of edge servers preclude the deployment of large-scale models. While smaller models, \textit{e.g.}, SDXL (3.5B) and Stable Diffusion 1.5 (0.9B), can be hosted, they incur prohibitive latency and artifacts when generating high-resolution images. Therefore, it remains challenging to provide high-resolution T2I generation service at the edge server without compromising latency.

Currently, super-resolution (SR) has emerged as a complementary technique to enhance the output quality of T2I models\cite{ho2022cascaded,ahmad2025alphaenhancer,zheng2024cogview3}, with some cascaded models like CogView3 \cite{zheng2024cogview3} integrating it through a multi-stage diffusion process. However, their reliance on sequential, large-scale diffusion architectures renders these methods computationally prohibitive for resource-constrained edge environments. Additionally, their fixed configurations, \textit{e.g.}, super-resolution scale, also hinder adaptation to dynamic requests and heterogeneous environment. The general SR methods themselves present a trade-off that lightweight learning-based approaches, \textit{e.g.}, Real-ESRGAN\cite{wang2021real}, SwinIR\cite{liang2021swinir}, enable fast inference but struggle to recover fine details, whereas diffusion-based SR, \textit{e.g.}, StableSR\cite{wang2024exploiting} and SinSR\cite{wang2024sinsr}, achieve superior detail recovery at the cost of high computational overhead. Moreover, a more fundamental limitation of exiting SR methods is their spatially invariant processing, which applies a uniform enhancement operation across the entire image, misaligned with the human visual system. Since human perception preferentially allocates attention to salient regions, such as main subjects, while being less sensitive to background detail, applying a one-size-fits-all enhancement strategy is both computationally inefficient and perceptually suboptimal\cite{HouTITS,tadin2019spatial}. 

To address these challenges, in this paper, we propose a collaborative generation-enhancement framework that synergizes the capabilities of both the edge and user devices. Our framework is based on a region-aware hybrid SR mechanism that applies distinct enhancement strategies based on content saliency. Upon receiving a T2I generation task, an edge-deployed T2I model first generates a low-resolution image based on adaptively selected denoising steps and super-resolution scales. This image is then spatially partitioned for parallel processing: the computationally intensive diffusion-based SR model on the edge server enhances the salient foreground region, while a lightweight SR model on the user device efficiently upscales the background area. This strategic allocation leverages the computational capacity of the edge for high-fidelity generative tasks, while offloading less demanding computation to the user device\cite{KongCSVT}. The enhanced foreground and background components are finally stitched on the user device to reconstruct the high-resolution image. This collaborative design achieves superior visual fidelity with reduced latency, effectively balancing the trade-off between perceptual quality and computational efficiency in resource-constrained edge environments. The main contributions of this paper are summarized as follows:
\vspace{-0.3em}
\begin{itemize}
\item  We propose an end-edge collaborative framework that coordinates generation and enhancement with an adaptive selection of denoising steps and super-resolution scale, tailored to dynamic requests and resource constraints.
\item We design a region-aware hybrid SR policy that applies diffusion-based SR to enhance salient regions for superior visual fidelity while utilizing efficient learning-based SR for less critical background regions, thereby balancing image quality and inference efficiency.
\item We evaluate the proposed framework through extensive experiments. Compared to other baselines, our framework achieves a 33\% reduction in service latency while maintaining competitive image quality.
\end{itemize}

\section{Motivation}
\begin{figure}[t]
\centering
    \vspace{-0.2cm}
    \subfigcapskip=-5pt
    \setlength{\abovecaptionskip}{-1pt}
     \subfigure[Delay]{
            \includegraphics[width=0.48\linewidth]{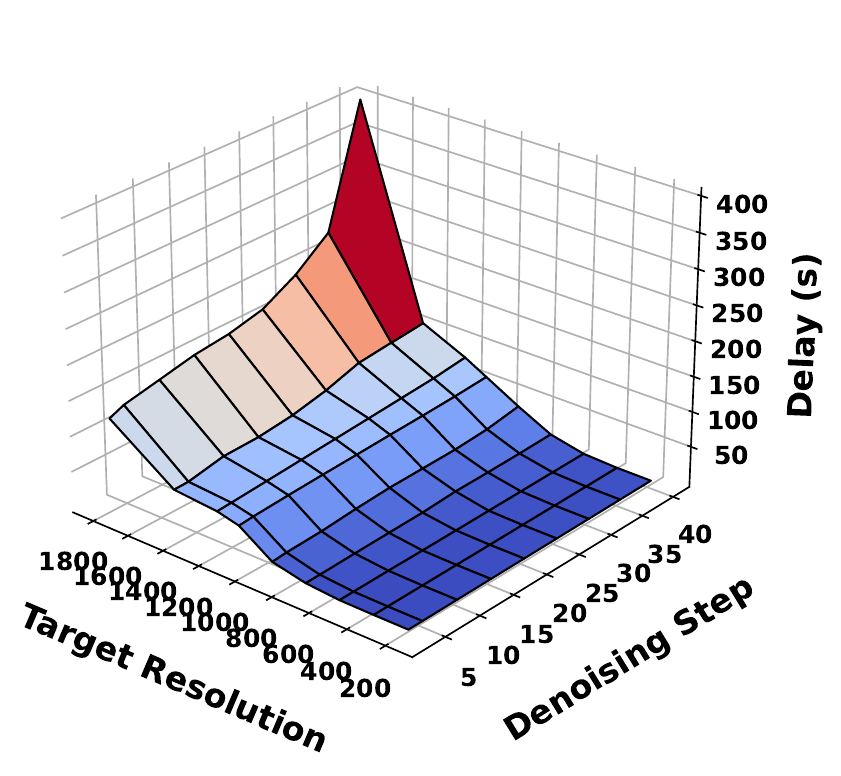}}
    \subfigure[Quality]{
            \includegraphics[width=0.48\linewidth]{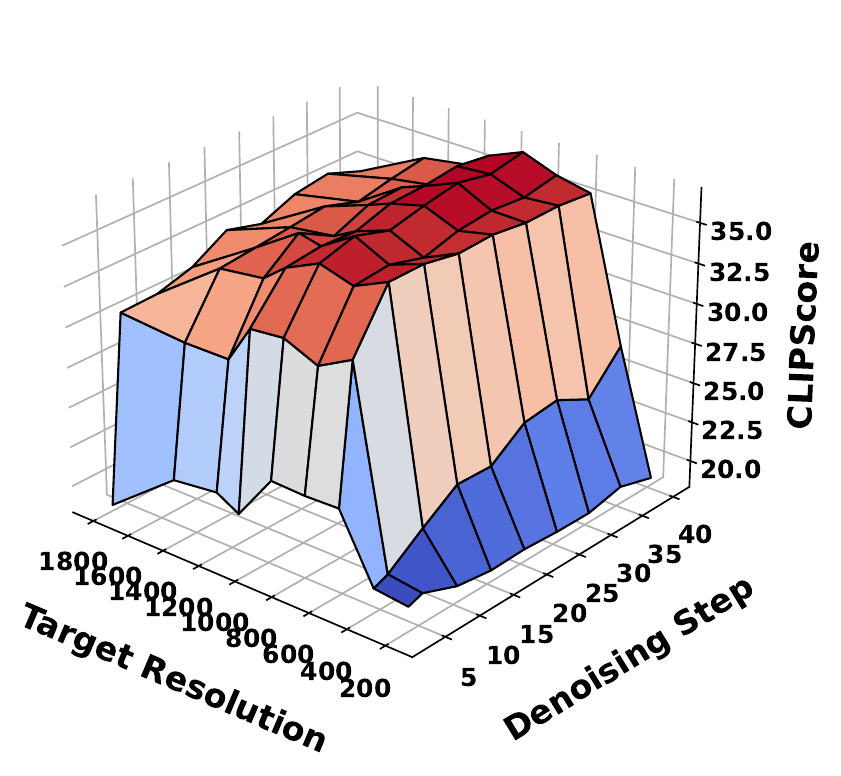}}
        \caption{The impact of SR scale and denoising steps. The horizontal axes indicate generation resolution (where 1000 denotes $1000\times1000$ pixels) and the number of denoising steps, respectively. }
    \label{fig:motivation1}
    \vspace{-0.3cm}
\end{figure}

\subsection{Quality-Latency Trade-Off in Edge T2I Generation }
Currently, high-resolution image generation has been a key focus of T2I models, as the level of resolution directly impacts the users' QoE. To characterize the limitations of direct high-resolution generation when deploying T2I models at the edge, we conduct a performance evaluation using the SDXL, a widely-adopted T2I model. As shown in Fig. \ref{fig:motivation1}, generation delay exhibits an exponential increase with respect to the target output resolution specified by users, while the image quality evaluated by CLIPScore\cite{hessel2021clipscore}, which assesses text-image alignment, initially improves before degrading due to artifacts and distortions. Furthermore, while delay scales linearly with the number of denoising steps, the corresponding quality improvement shows diminishing marginal returns. 

\begin{figure}[t]
\centering
    \setlength{\abovecaptionskip}{-1pt}
     \subfigure[Relatively lower target resolutions]{
            \includegraphics[width=0.48\linewidth]{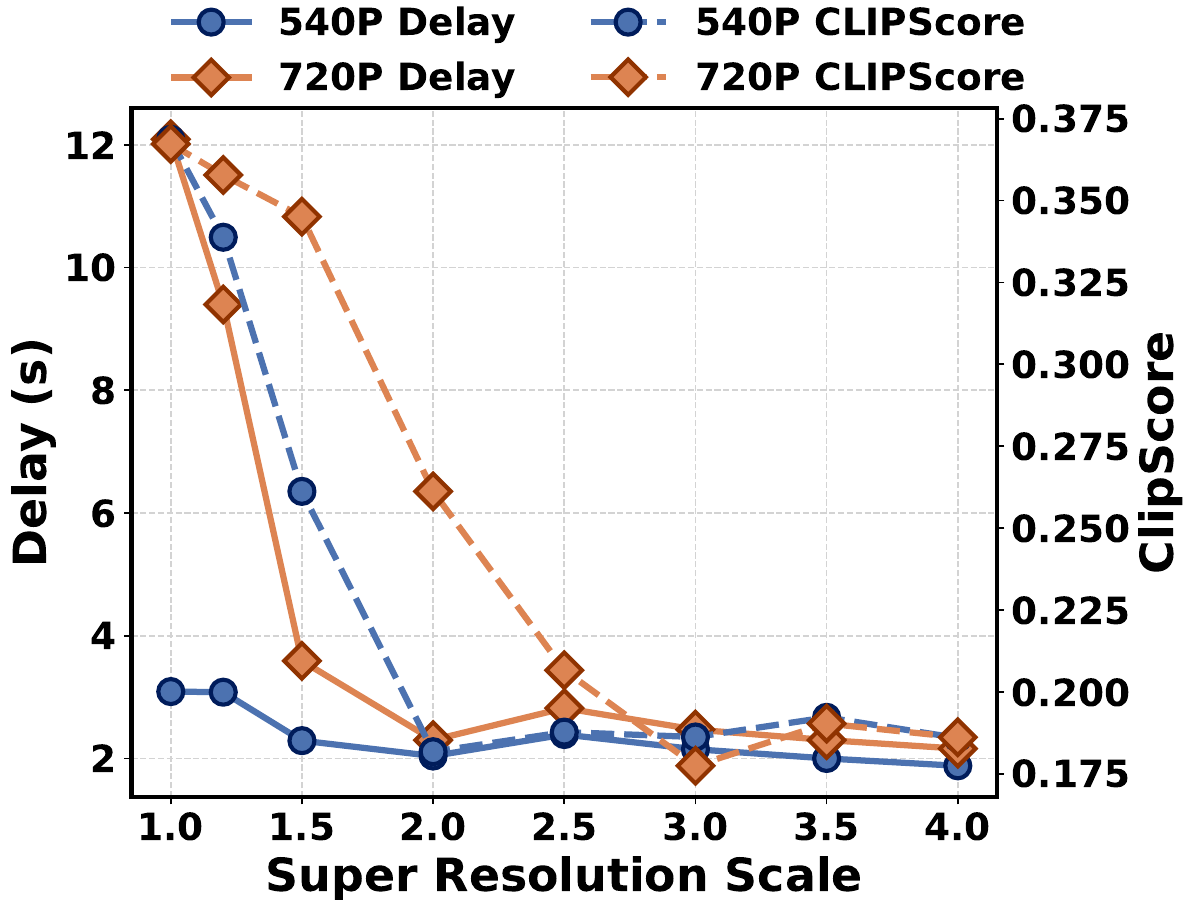}}
    \subfigure[Relatively higher target resolutions]{
            \includegraphics[width=0.48\linewidth]{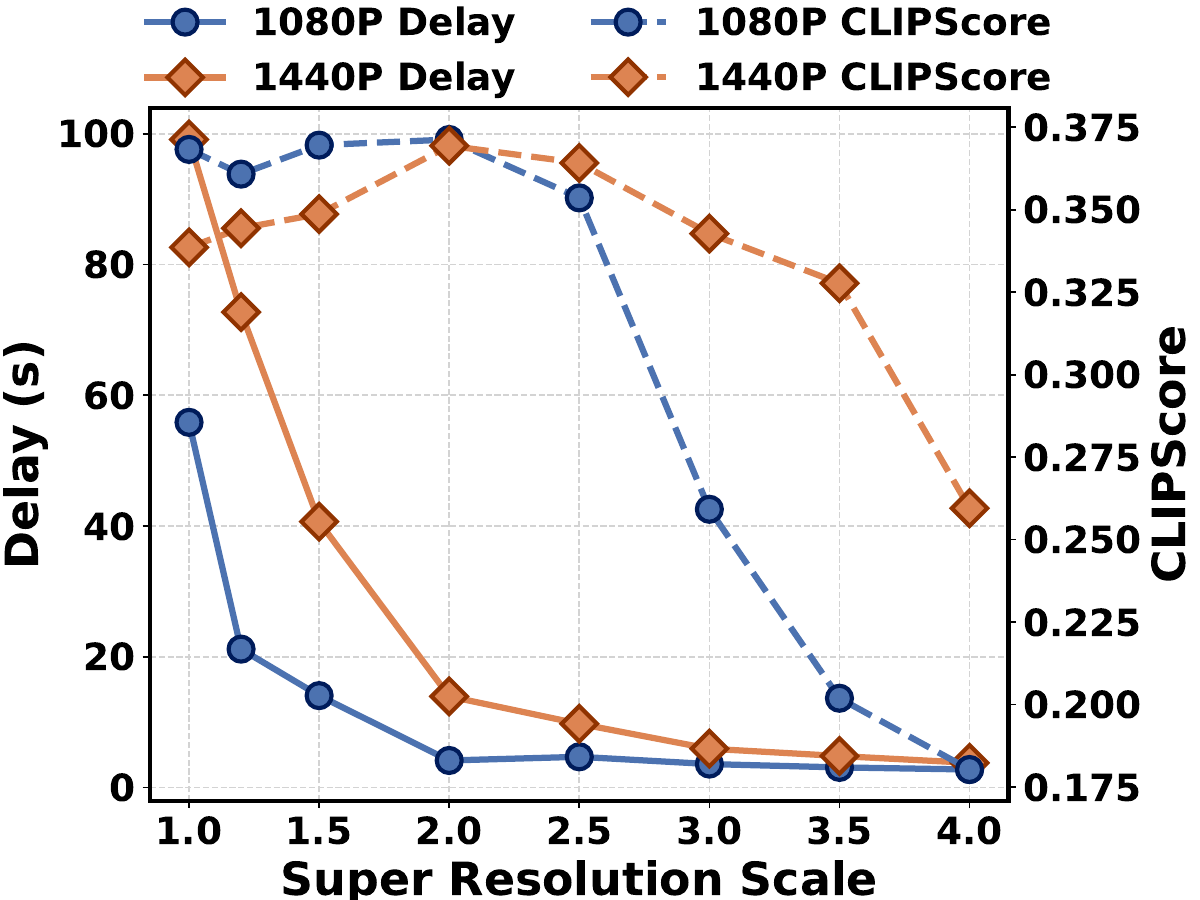}}
        \caption{Generation delay and Clipscore for different target resolutions.}
    \label{fig:motivation2}
    \vspace{-0.3cm}
\end{figure}

Stemming from these limitations of generating high-resolution images directly,  we explore the feasibility of a collaborative generation-enhancement framework\cite{ZhouTMCUser,zhou2025user_wcm}, which leverages SR for subsequent image reconstruction.
We evaluate this framework using various SR scale settings for different target resolutions. As shown in Fig. \ref{fig:motivation2}, selecting an appropriate SR scale for the target resolution effectively reduces service latency while preserving high image quality. Specifically, for relatively lower target resolutions, \textit{e.g.}, $540P$ and $720P$ in Fig. \ref{fig:motivation2}(a), direct generation or a smaller SR scale strikes a good balance between quality and responsiveness. However, for relatively higher target resolutions, \textit{e.g.}, $1080P$ and $1440P$ in Fig. \ref{fig:motivation2}(b), a larger SR scale significantly reduces latency while ensuring comparable image fidelity.
Our experiments validate the collaborative generation-enhancement framework as an effective solution for high-resolution T2I generation, which necessitates an adaptive controller to dynamically configure key parameters, \textit{e.g.}, super-resolution scale and denoising steps, based on the target output resolution.

\subsection{Region-aware Hybrid Super-Resolution Policy}
\begin{figure}[t]
\centering
    \setlength{\abovecaptionskip}{-1pt}
    \subfigure[Learning-based SR]{
        \includegraphics[width=0.3\linewidth]{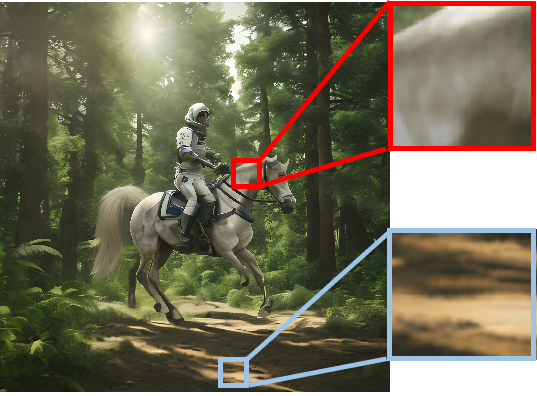}
    }\hfill
    \subfigure[Diffusion-based SR]{
        \includegraphics[width=0.3\linewidth]{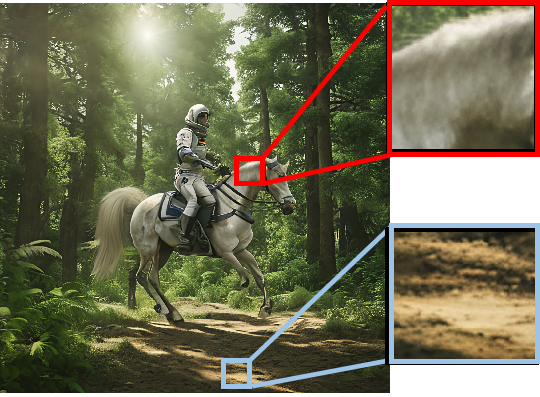}
    }\hfill
    \subfigure[Hybrid SR]{
        \includegraphics[width=0.3\linewidth]{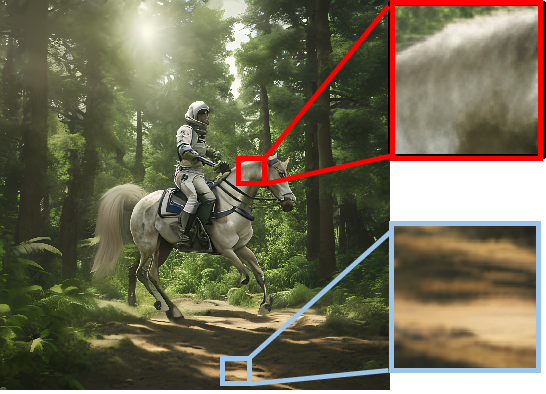}
    }
    \caption{Visual comparison among different super-resolution strategies.}
    \label{fig:motivation3}
    \vspace{-0.3cm}
\end{figure}

While the collaborative framework is functional, its final performance is critically dependent on the SR module. To this end, we experimentally evaluated two mainstream categories of SR methods, \textit{i.e.}, learning-based and diffusion-based models, by employing both as enhancement modules for the low-resolution images generated by the T2I model. As shown in Fig. \ref{fig:motivation3}(a) and (b), while the learning-based SR model (16.7M) requires only $8.41\text{s}$ on average to perform $4\times$ upscaling on a $512 \times 512$ image, it struggles to recover fine-grained details, often producing overly smooth results. In contrast, the diffusion-based SR model (1B) achieves superior detail reconstruction due to its inherent generative capability, but at a high computational cost of $125.4\text{s}$ for the same task.

To resolve this conflict, we observe that the human visual system is more sensitive to fine details in salient regions, such as human faces or key objects, than in background areas. This insight motivates a region-aware hybrid super-resolution policy that applying diffusion-based SR to visually important foreground regions to ensure high-fidelity detail recovery, while coordinating lightweight learning-based SR for efficient upscaling of background areas. As shown in Fig. \ref{fig:motivation3}(c), this hybrid approach achieves significantly improved perceptual quality with only a marginal latency increase of $8.3\text{s}$, effectively bridging the gap between detail fidelity and computational efficiency. 
These findings underscore the value of the region-aware hybrid SR policy that strategically allocates high-fidelity SR to salient regions while employing efficient upscaling for backgrounds, thereby achieving an optimal balance between visual quality and practical performance. 
\section{System Model and Problem Formulation}
\subsection{System Model}
\subsubsection{Region-Aware Latent Partitioner}

\begin{figure}[t]
\centering
    \setlength{\abovecaptionskip}{-1pt}
    \subfigure[Case1: Patches]{
        \includegraphics[width=0.22\linewidth]{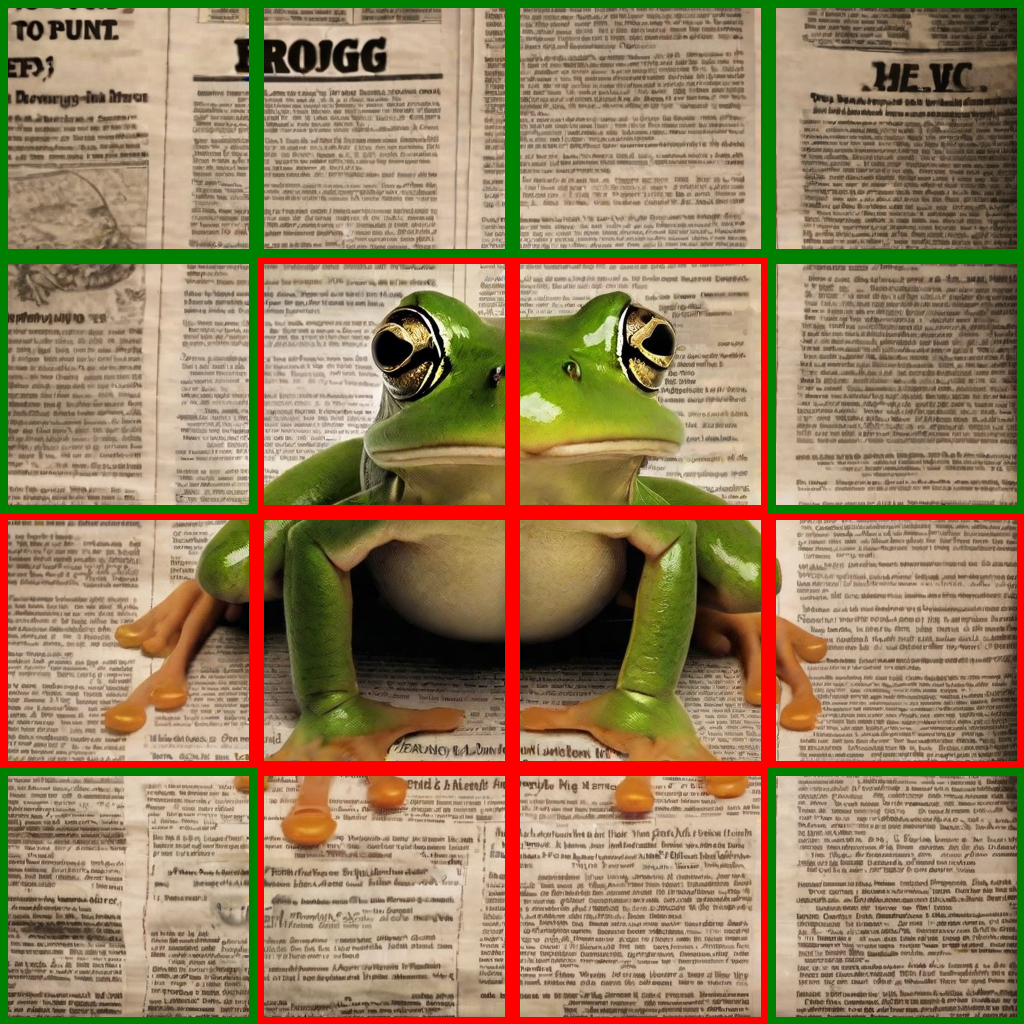}
    }\hfill
    \subfigure[Case1: SAM]{
        \includegraphics[width=0.22\linewidth]{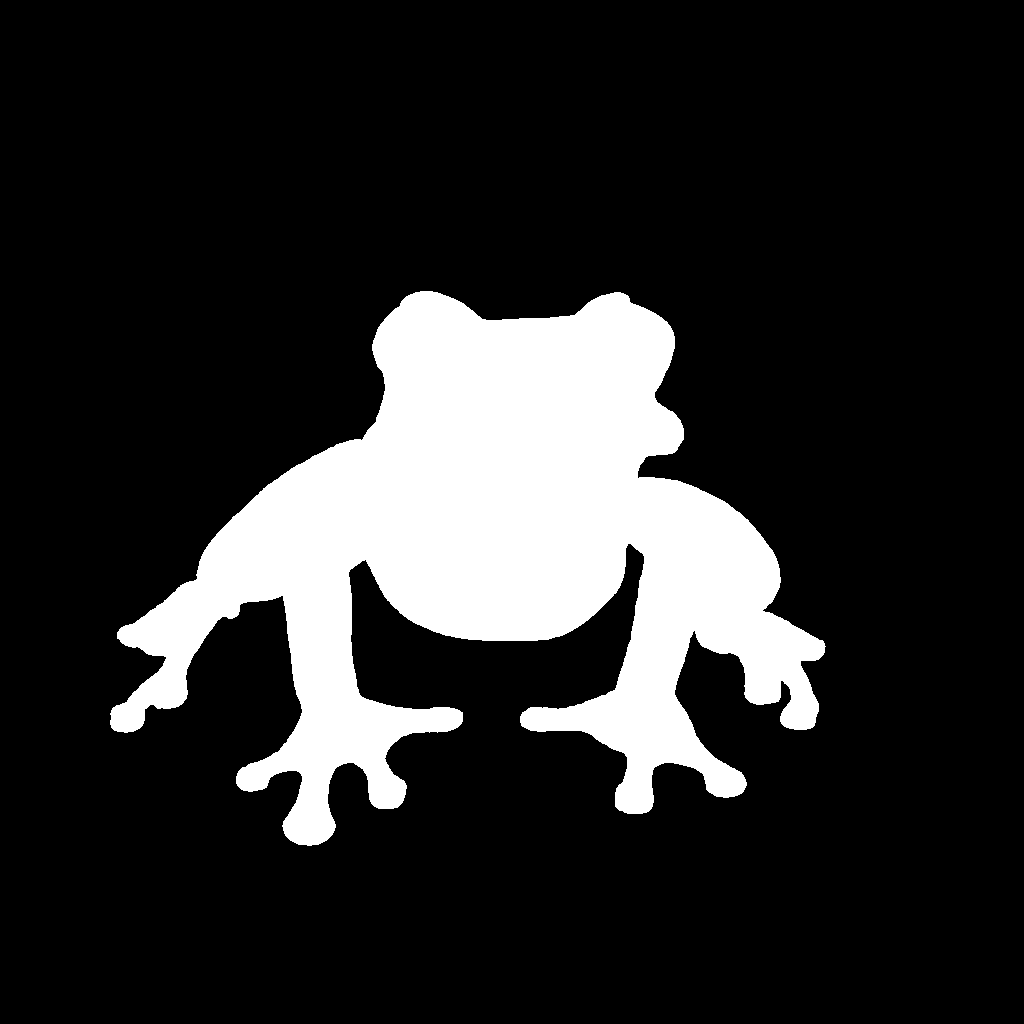}
    }\hfill
    \subfigure[Case2: Patches]{
        \includegraphics[width=0.22\linewidth]{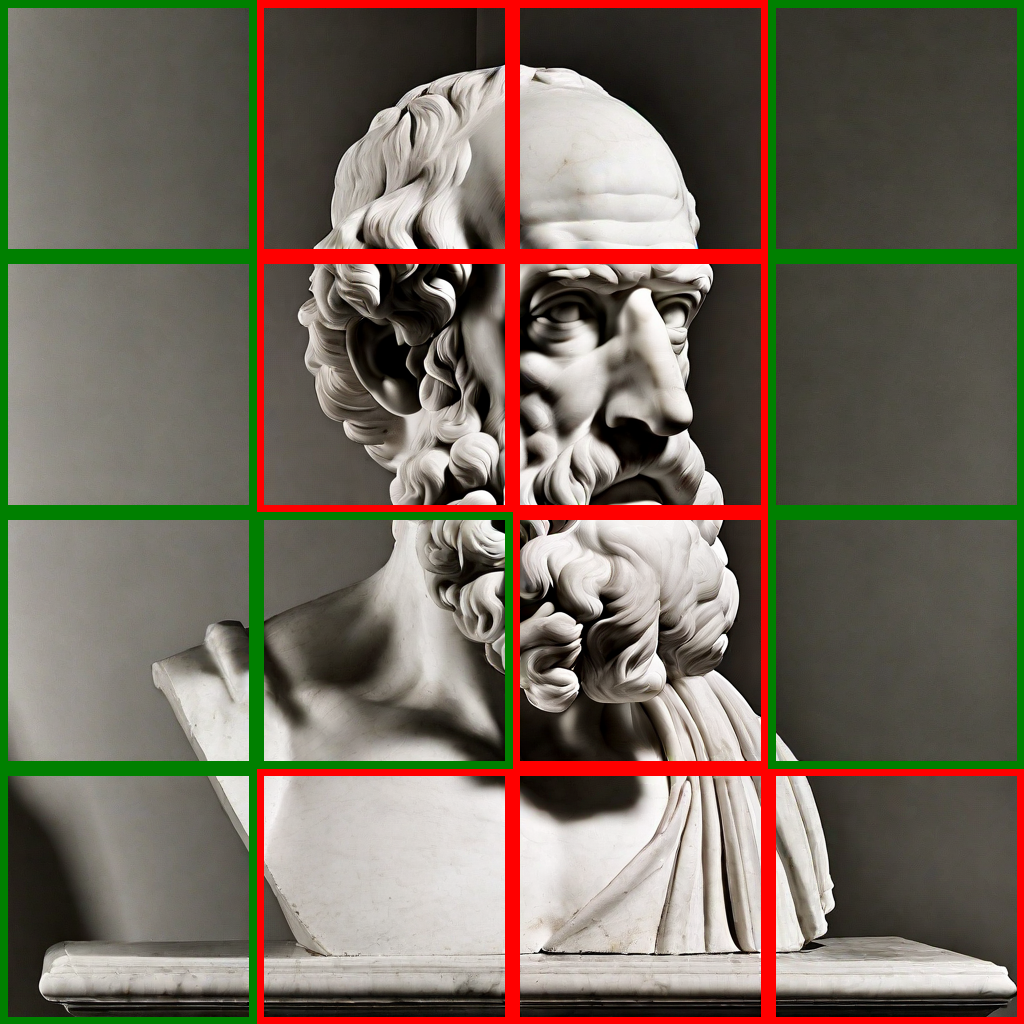}
    }\hfill
    \subfigure[Case2: SAM]{
        \includegraphics[width=0.22\linewidth]{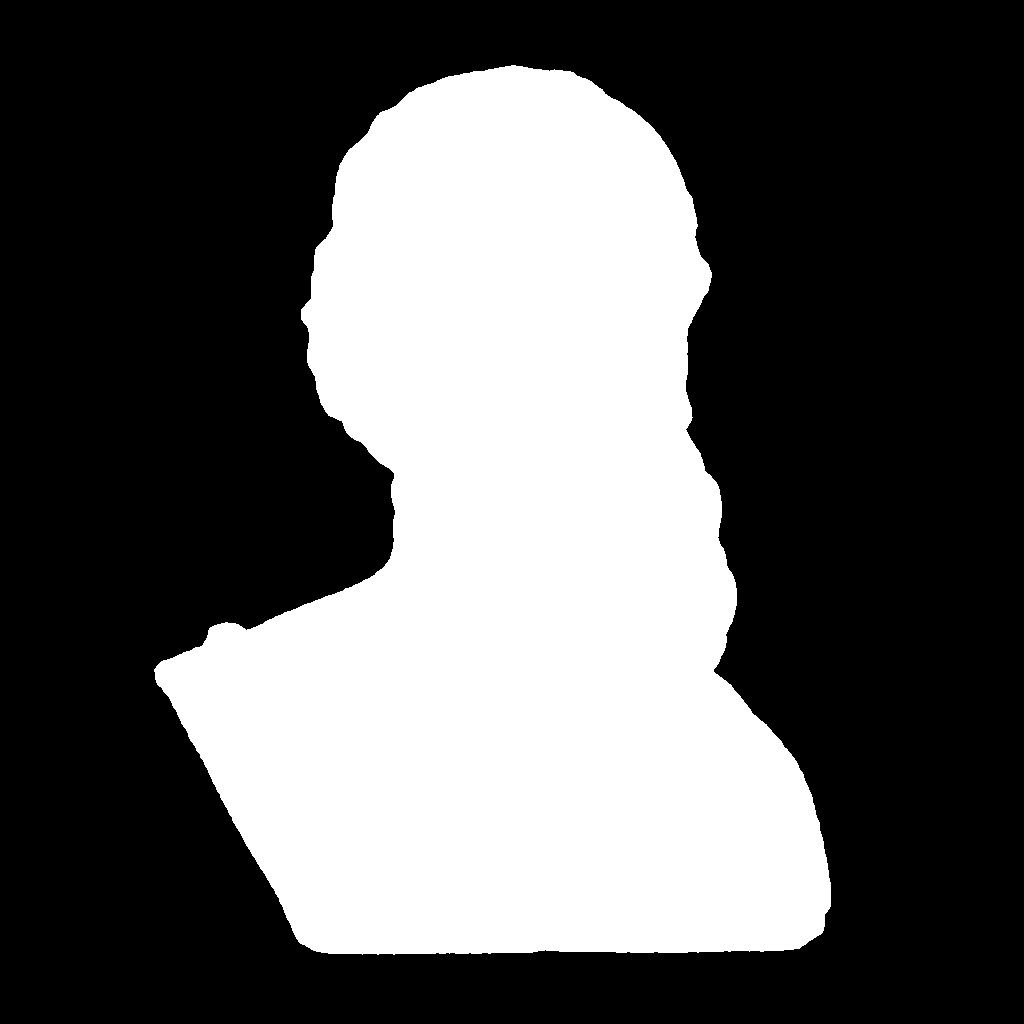}
    }
    \caption{Visualization of variance-based latent partitioning, where red/green patches represent high/low variance, and white/black SAM areas represent foreground/background regions.}
    \label{fig:compare with SAM}
    \vspace{-0.3cm}
\end{figure}

We develop a latent partitioner built on the mentioned region-aware hybrid SR policy, eliminating reliance on external semantic segmentation models. The effectiveness of this lightweight partitioner is empirically validated against the Segment Anything Model (SAM)\cite{kirillov2023segment}, a powerful segmentation model. Specifically, latent features are partitioned into a $4\times4$ grid, with the top half by spatial variance designated as high-variance and the remainder as low-variance. Fig. \ref{fig:compare with SAM} illustrates the grid and the corresponding SAM masks. High-variance patches exhibit strong correspondence with SAM-identified salient regions, achieving a mean IoU of 0.7706, while low-variance patches align predominantly with background, validating that spatial variance in the latent space effectively distinguishes foreground from background.

\subsubsection{System Overview}
\begin{figure}[t]
\subfigcapskip=-2pt
    \centering
    \includegraphics[width=1\linewidth]{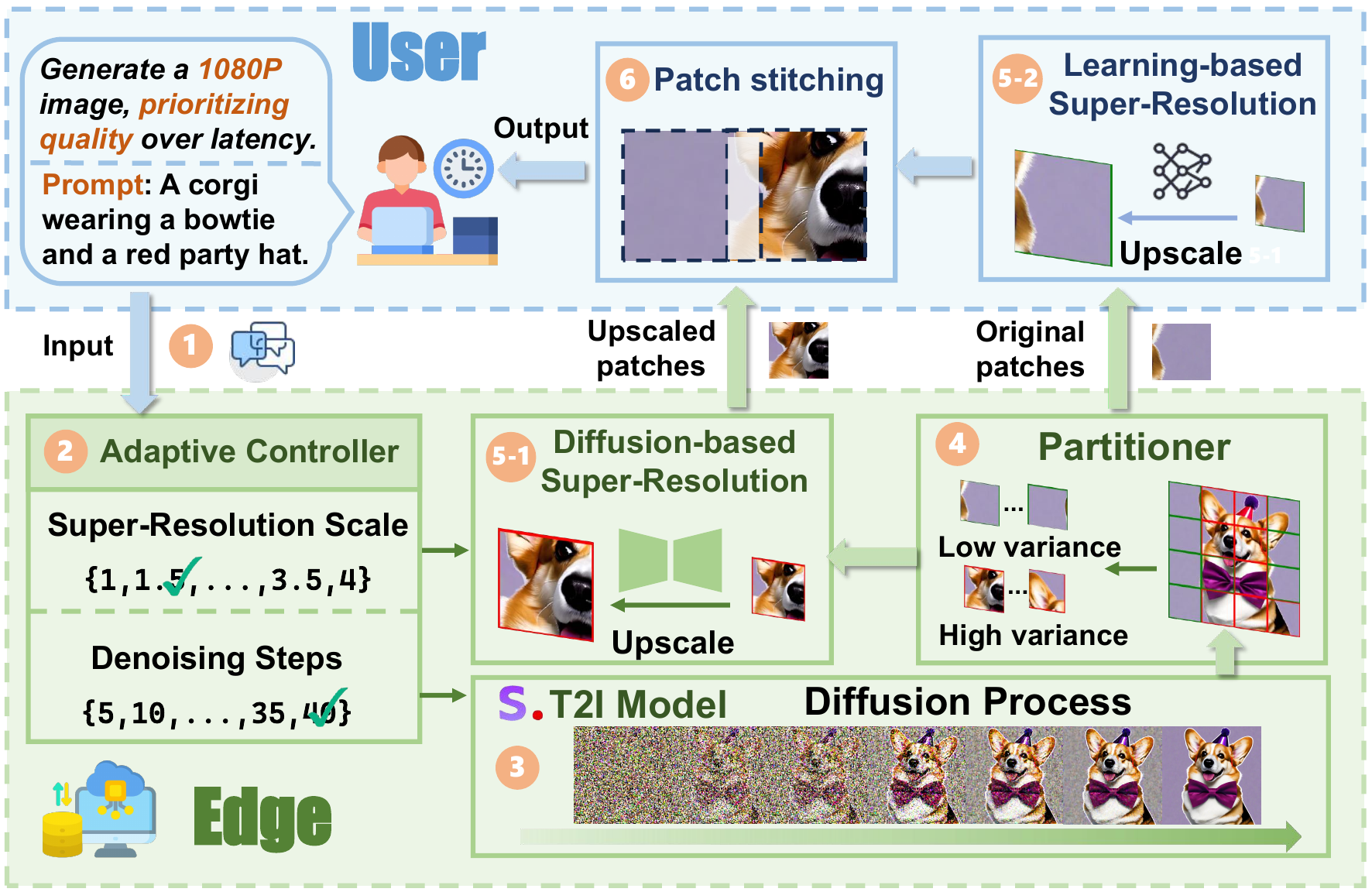}
    \caption{System Overview}
    \label{fig:placeholder}
    \vspace{-0.3cm}
\end{figure}

As shown in Fig. \ref{fig:placeholder}, we design an end-edge collaborative generation-enhancement system to provide T2I generation services for a set of users $\mathcal{K}=\{1,\dots,K\}$. 
On receiving a generation task $q_k=\{\tilde{R}_k,\lambda_k,p_k\}$ from user  $k$, where $\tilde{R}_k$ denotes the target resolution, $\lambda_k\in[0,1]$ represents the preference weight between image quality and latency, and $p_k$ is the textual prompt, the adaptive controller determines the operational configuration  $\boldsymbol{e}_k=[S_k,D_k]$ for this task, where $S_k$ is the SR scale selected from a finite candidate set $\mathcal{S}=\{q^{1}, \dots, q^S\}$, and $D_k$ is the number of denoising steps chosen from the set $\mathcal{D}=\{g^{1}, \dots, g^D\}$ . The initial generation resolution is thus determined by the target resolution $\tilde{R}_k$ and the selected SR scale, \textit{i.e.}, $R_k = \frac{\tilde{R}_k}{S_k}$.

Following the initial generation of the latent feature by the diffusion-based T2I model, a region-aware partitioner segments it based on spatial variance before it is decode into the image. Specifically, the latent feature is divided into multiple patches, where higher-variance regions form the foreground set $\Omega_k^{\mathrm{1}}$ and lower-variance regions constitute the background set $\Omega_k^{\mathrm{2}}$. The corresponding patches are enhanced in parallel that the diffusion-based SR model refines $\Omega_k^{\mathrm{1}}$ to recover texture and structure at the edge, while the user-side learning-based SR model upscales $\Omega_k^{\mathrm{2}}$ efficiently. Finally, on the user device, all enhanced patches are stitched with feathered overlap–add, using weighted overlaps to suppress blocking artifacts and yield the final high-resolution image $\tilde{I}_k$.

\subsubsection{Image Quality Model}\label{B}

Diffusion-based text-to-image generation progressively synthesizes images by iteratively denoising Gaussian noise under text guidance. As shown in Fig. \ref{fig:motivation1}(b), both the denoising steps $D_k$ and initial resolution $R_k$ critically influence output quality which can be expressed as: 
\begin{equation}
Q_k^1 = g(D_k, R_k).
\label{eq:Qk}
\end{equation}

The allocation ratio $\gamma \in [0, 1]$, defined as the fraction of image patches processed by the diffusion-based SR model, also influences the enhancement outcome. 
A higher value of $\gamma$ improves quality by engaging the detail restoration capability of the diffusion model across more patches.
Therefore, we define the final generation quality of the request $k$ as:
\begin{equation}
    Q_{k} = Q_k^{1} \cdot m(\gamma),
\end{equation}
where $m(\gamma) > 1$ represents the improvement ratio of image quality with respect to the parameter $\gamma$.

\subsubsection{Latency Model}\label{C}
For the T2I generation task $q_k$, its service latency primarily comprises inference, super-resolution and transmission latency.
The inference latency of the initial low-resolution image generation at the edge is determined by the number of denoising steps and the initial generation resolution according to Fig.~\ref{fig:motivation1}(b) and is expressed as:
\begin{equation}
    T_k^{\mathrm{1}} = \frac{\xi_k^{\mathrm{1}}(D_k, R_k)}{C_k^{\mathrm{1}}},
\label{eq:Tgen_load}
\end{equation}
where $\xi_k^{\mathrm{1}}(D_k, R_k)$ denotes the computation load determined by the number of denoising steps $D_k$ and the initial resolution $R_k$, and $C_k^{\mathrm{1}}$ is the available computing capacity at the edge.

After the latent features of the initial low-resolution image are generated, the region-aware latent partitioner divides them into the foreground region set $\Omega_k^{\mathrm{1}}$ and background region set $\Omega_k^{\mathrm{2}}$ based on variances with negligible delay. The former are enhanced by the diffusion-based SR model deployed at the edge, while the latter are processed by the learning-based SR model on the user device. According to \cite{wang2020deep}, the processing latency for each SR branch primarily depends on the initial resolution. Thus, the super-resolution latency at the edge and the user side are respectively expressed as:
\begin{equation}
    T_k^{\mathrm{2}} = \frac{\xi_k^{\mathrm{2}}(R_k,\gamma)}{C_k^{\mathrm{1}}},
    \quad
    T_k^{\mathrm{3}} = \frac{\xi_k^{\mathrm{3}}(R_k,\gamma)}{C_k^{\mathrm{2}}},
\label{eq:TSR_load}
\end{equation}
where $\xi_k^{\mathrm{2}}(R_k,\gamma)$ and $\xi_k^{\mathrm{3}}(R_k,\gamma)$ denote the computation loads associated with the the initial resolution $R_k$ and allocation ratio $\gamma$. And $C_k^{\mathrm{1}}$ and $C_k^{\mathrm{2}}$ represent the available computing capacity on the edge server and user device, respectively.




Within the hybrid enhancement pipeline, two parallel transmission paths are established between the edge server and the user device. The first path is dedicated to high-variance patches, which are processed by the diffusion-based SR model deployed at the edge. Subsequently, the upscaled patches are transmitted to the user side for stitching, incurring a transmission latency $T_k^{\mathrm{4}}$. Concurrently, the second path handles the low-variance patches, which are transmitted directly from the edge to the user side for local SR via learning-based SR model, leading to the transmission latency $T_k^{\mathrm{5}}$. Both latency components are determined by the data volume and the available network bandwidth. We formulate them as follows:
\begin{equation}
    T_k^{\mathrm{4}} = \frac{\eta_k(\tilde{R}_k,\gamma)}{B_k},
     \quad
         T_k^{\mathrm{5}} = \frac{\eta_k(R_k,\gamma)}{B_k},
\end{equation}
where $B_k$ is the transmission bandwidth between the edge and the user $k$, and $\eta_k(\cdot,\gamma)$ represents the transmitted data volume, which is contingent on the allocation ratio $\gamma$. Notably, for $T_k^{\mathrm{4}}$, the data volume depends on the target resolution $\tilde{R}_k$ of the upscaled patches, whereas for $T_k^{\mathrm{5}}$, it depends on the initial resolution $R_k$ of the patches being transmitted directly.




As the collaborative enhancement process is parallel for both SR branches, the total enhancement latency $T_k^{\mathrm{6}}$ is governed by the slower of the two parallel workflows, which can be expressed as follows:
\begin{equation}
    T_k^{\mathrm{6}} = \max\!\big(T_k^{\mathrm{2}} + T_k^{\mathrm{4}},\; T_k^{\mathrm{3}} + T_k^{\mathrm{5}}\big).
\label{eq:TSR_parallel}
\end{equation}

The service latency for the task $q_k$ can be formulated as:
\begin{equation}
    T_k = T_k^{\mathrm{1}} + T_k^{\mathrm{6}},
\label{eq:Tk_total_load}
\end{equation}
which comprehensively captures the time elapsed from receiving a task to the final delivery of the enhanced result. The allocation ratio $\gamma$ is fixed at 0.25 in our implementation as detailed in Section IV.

\subsection{Problem Formulation}\label{D}
The objective of the proposed end-edge collaborative system is to achieve the trade-off between the image generation quality and the service latency. To this end, we formulate a utility function for the task $q_k$ to quantify this balance: 
\begin{equation}
    U_k = Q_k - \lambda_k T_k,
\label{eq:Uk_final}
\end{equation}
where $Q_k$ and $T_k$ represent the image quality and the service latency, respectively. $\lambda_k$ is a user-specific weight that balances preference for image quality against latency. A lower value of $\lambda_k$ signifies a greater emphasis on image fidelity, whereas a higher value prioritizes service latency.
\begin{algorithm}[t]
\caption{Configuration Selection Algorithm}
\label{alg:SA}
\begin{algorithmic}[1]
\Require Initial configuration $e_k=[S_k,D_k]$, utility function $U(e_k)$, temperature decay coefficient $\alpha$, initial temperature $T$, minimum temperature $T_{\mathrm{min}}$, iteration limit per temperature $\mathrm{limit}_T$
\State Initialize temperature $T$ and cooling coefficient $\alpha$
\State Compute utility $U(e_k)$
\While{$T > T_{\mathrm{min}}$}
    \For{$t = 1$ to $\mathrm{limit}_T$}
        \State Generate neighboring configuration $e_k'$ of $e_k$
        \If{$\mathrm{GetLatency}(e_k') \leq L_0$}
            \State Compute utility $U(e'_k)$
            \State $\Delta U \gets U(e'_k) - U(e_k)$
            \State Accept $e_k' \gets e_k$ if $\Delta U > 0$ 
            \State Else accept with probability $\exp(-\Delta U / T)$
        \EndIf
    \EndFor
    \State Update temperature: $T \gets T \times \alpha$
\EndWhile
\State \Return $e_k$
\end{algorithmic}
\end{algorithm}
Our system aims to maximize the utility across all concurrent users by determining the optimal configuration $e_k=[S_k, D_k]$ for each task. The optimization problem can thus be formulated as:
\begin{align}
\mathbf{P}:&\max_{\{\boldsymbol{e}_k=[S_k, D_k]\}} \sum_{k \in \mathcal{K}} U_k,
\label{eq:P_obj_final}\\
\text{s.t.}\qquad\qquad & S_k \in \mathcal{S},D_k \in \mathcal{D} \quad \forall k \in \mathcal{K}, \label{eq:P_C1_final}\\
 & \sum_{k \in \mathcal{K}} \xi_k^{\mathrm{1}}(D_k,R_k) + \xi_k^{\mathrm{2}}(R_k,\gamma) \leq C_{\max}, \label{eq:P_C5_final}\\
 & \sum_{k \in \mathcal{K}} \xi_k^{\mathrm{3}}(R_k,\gamma)  \leq C^{\prime}_{\max},\label{eq:P_C6_final}\\
 & R_k=\frac{\tilde{R}_k}{S_k}, \quad \forall k \in \mathcal{K}, \label{eq:P_C7_final}
\end{align}
where Constraint~(\ref{eq:P_C1_final}) specify the feasible range of the SR scale and the number of denoising steps. Constraints~(\ref{eq:P_C5_final}) and~(\ref{eq:P_C6_final}) enforce the computational resource limits of the edge server and the user device, denoted by $C_{\max}$ and $C^{\prime}_{\max}$, respectively. Constraint~(\ref{eq:P_C7_final}) establishes the relationship between the initial generation resolution and selected SR scale.
\subsection{Algorithm Design}\label{E}

The optimization problem formulated in $\mathbf{P}$ is combinatorial, making an exhaustive search for the optimal configuration $e_k=[S_k,D_k]$ computationally intractable, particularly given the dynamic nature of user requests and system resource availability in edge environments. To address this challenge, we develop an efficient configuration selection algorithm based on Simulated Annealing (SA), as detailed in Algorithm \ref{alg:SA}. The algorithm begins by initializing the system temperature $T$ and cooling rate $\alpha$. At each temperature stage, it performs multiple iterations to thoroughly explore the solution space. In each iteration, the current configuration is perturbed to generate a neighboring solution, which is evaluated for utility improvement. The acceptance criterion follows the Metropolis rule that configurations with enhanced utility are accepted directly, while those with degraded performance may still be accepted with probability $\exp(-\Delta U / T)$, thus effectively balancing exploration and exploitation throughout the optimization process. This probabilistic acceptance mechanism allows the algorithm to escape local optima in early stages while converging toward high-quality solutions in later phases. The outer loop systematically reduces the temperature according to $T \gets T \times \alpha$, while the inner loop conducts extensive searches at each temperature level. Through this carefully designed annealing schedule, the algorithm efficiently navigates the complex solution space and returns a near-optimal configuration that effectively balances image quality and service latency in the collaborative T2I generation framework.

\section{Performance Evaluation}\label{F}
\subsection{Experiment Setup}
\subsubsection{Implementation}
We consider an edge server serving T2I generation requests for ten users ($K=10$). The edge server is equipped with an NVIDIA GeForce RTX 3080 Ti (34.1 TFLOPS). We emulate the user device with an NVIDIA Jetson AGX Orin (10.6 TFLOPS). The edge–device link bandwidth is fixed at 10 Mbps. In addition, SDXL, StableSR and Real-ESRGAN are used as the edge-deployed T2I model, edge side diffusion-based SR model, and user side learning-based SR model, respectively, owing to their demonstrable performance and broad adopted in related research.
\subsubsection{Baseline} We compare our proposed system with the following baselines:
\begin{itemize}
    \item \textbf{Random}: This method randomly selects different initial generation resolutions and denoising steps.
    \item \textbf{w/o SR}: This method does not use super-resolution enhancement technology, but directly generates the image with the target resolution.
    \item \textbf{OneType}: This method tunes only the denoising step count, while the super-resolution factor defaults to $2\times$.
    \item \textbf{Congview3\cite{zheng2024cogview3}}: CogView3 is a cascaded text-to-image diffusion framework which first synthesizes a low-resolution draft, then progressively upsamples and refines it via relay super-resolution diffusion.

\end{itemize}
\begin{figure}[t]
    \vspace{-0.2cm}
    \centering
    \includegraphics[width=1.0\linewidth]{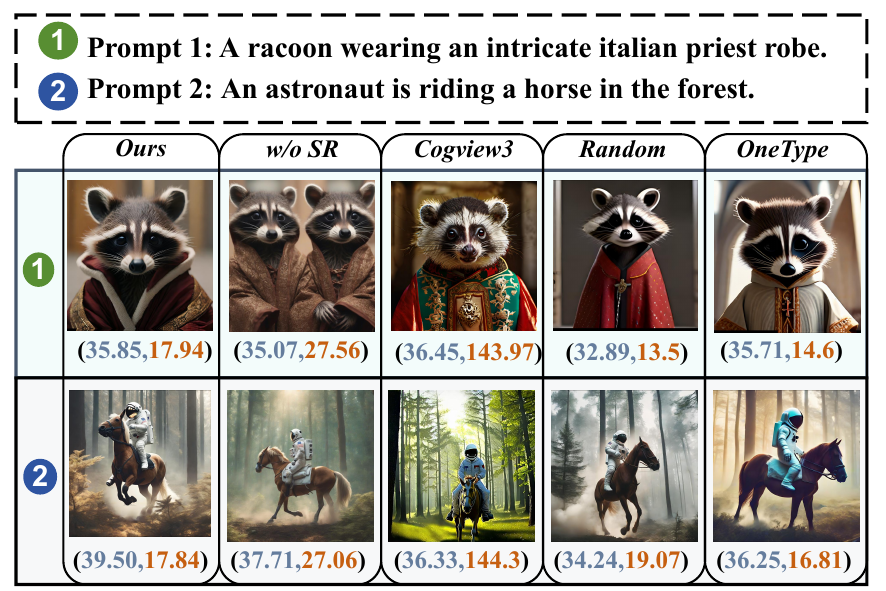}
    \caption{Visualization of each method with the target resolution $1024\times1024$, where $(x, y)$ denotes the CLIPScore and the service latency (s), respectively.}
    \label{fig:Visualization}
    \vspace{-0.3cm}
\end{figure}

\begin{figure*}[t]
\centering
    \subfigcapskip=-3pt
    \setlength{\abovecaptionskip}{-1pt}
     \subfigure[The trade-off between service latency and quality at different target resolutions]{
            \includegraphics[width=0.48\linewidth]{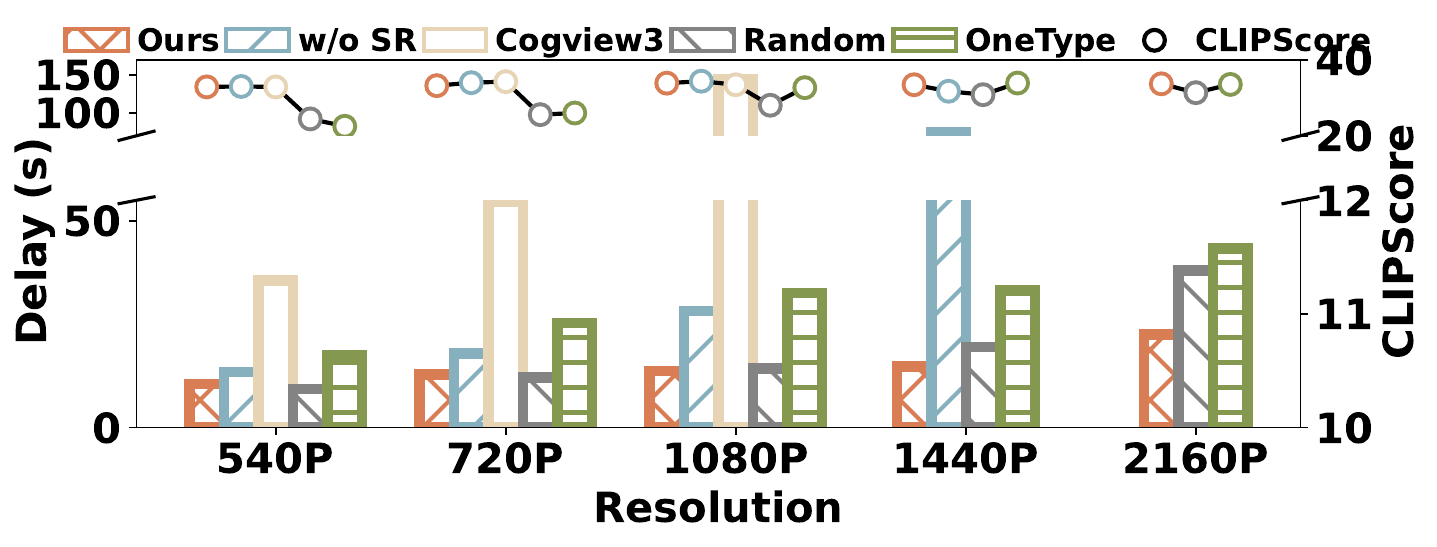}}
    \subfigure[The comparison of utility at different target resolutions]{
            \includegraphics[width=0.47\linewidth]{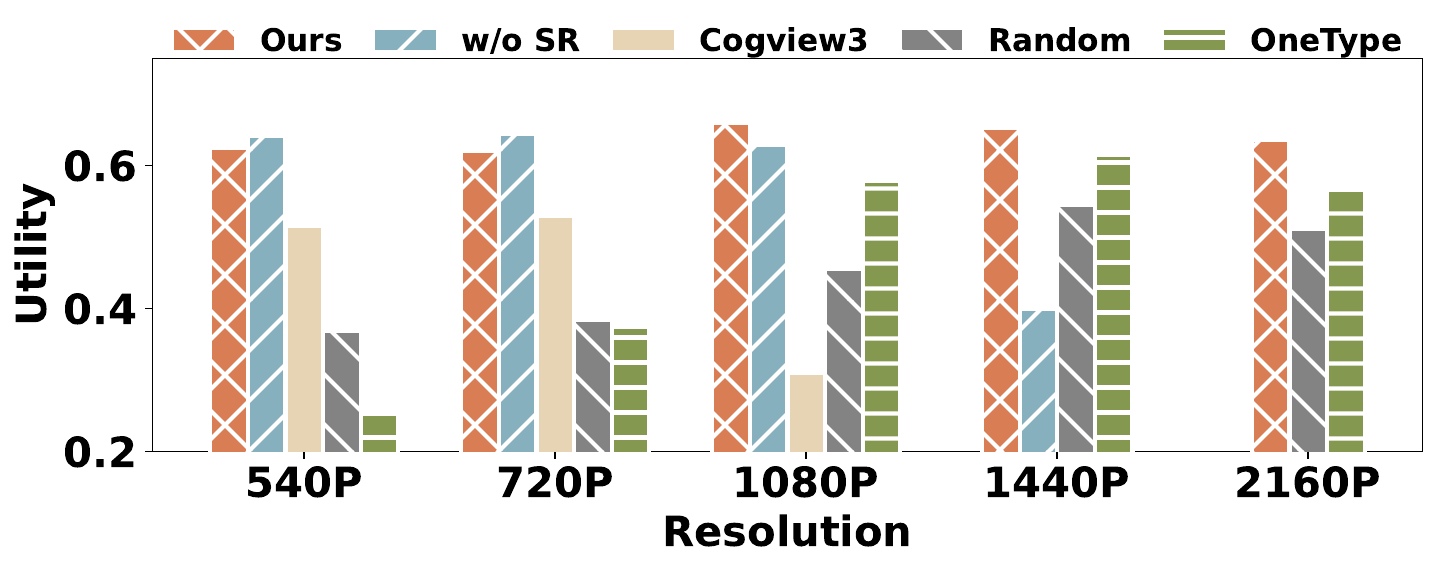}}

        \caption{The performance of each method across different target resolutions.}
    \label{fig:result1}
    \vspace{-0.3cm}
\end{figure*}

\subsection{Quality-Latency Trade-Off}


We begin with a visual comparison of different methods at the same target resolution in Fig. \ref{fig:Visualization}. In the first image generation task, our method offers an exceptional trade-off with only a marginal 1.6\% deficit in image quality compared to CogView3 while achieving an $8\times$ inference speedup. Additionally, it delivers up to an 8.9\% quality improvement over other methods at a comparable latency. In the second task, our method surpasses all baselines in generation quality and attains the second-highest inference speed.


We then compare the proposed method with the baselines under a comprehensive evaluation scenario using P2\footnote{https://huggingface.co/datasets/nateraw/parti-prompts} dataset. The comparison is conducted along three key dimensions: image quality, service latency, and overall utility.
As shown in Fig. \ref{fig:result1}(a), at lower target resolutions \textit{e.g.}, $720P$, our method delivers slightly lower image quality than the w/o SR and CogView3 variants by 1.5\% and 1.2\%, respectively, but achieves the lowest service latency, reducing latency by 16\% and 70\%, respectively. At $1080P$, our approach maintains nearly stable image quality with only a modest average latency increase of $1.6\text{s}$. In contrast, the w/o SR method exhibits a significant $7.8\text{s}$ latency increase, while CogView3 suffers from both a slight quality drop and a $2\times$ slowdown. Notably, CogView3's performance degrades markedly beyond $1080P$, with steep latency growth making $1440P$ and $2K$ generation infeasible. Additionally, our method outperforms both Random and OneType in speed and quality on average. Consequently, it ensures stable quality while reducing latency by 33\% compared to other methods.

Fig. \ref{fig:result1}(b) compares the overall utility across all methods. Our method maintains stable utility across all resolutions. Although w/o SR marginally surpasses ours at low resolutions, its utility drops sharply at higher resolutions and does not support $2K$ generation. The utility of the OneType method improves with resolution but fails to reach the optimum due to a lack of precise control. Overall, our method yields an average utility improvement of 25\% over the baselines.

\subsection{Ablation Study}
\subsubsection{Impact of Edge Computing Capacity}
We then compare the average overall utility of our approach with baselines under varying levels of edge computing capacity. As shown in Fig. \ref{fig:result2}(a), the x-axis denotes the edge compute availability ratio, \textit{i.e.}, the fraction of the default compute resources used in our main setup. As the available compute decreases, the proposed method consistently attains higher utility than all alternatives and exhibits notably smaller fluctuations, whereas the baselines show an approximately linear decline. This advantage stems from our adaptive configuration strategy, which reallocates steps and SR choices to preserve a favorable balance between generation quality and latency across heterogeneous environments.
\subsubsection{Impact of Patch Allocation Ratio}
\begin{figure}[t]
\centering
    \vspace{-0.1cm}
    \subfigcapskip=-3pt
    \setlength{\abovecaptionskip}{-1pt}
         \subfigure[The edge computing capacity]{
            \includegraphics[width=0.47\linewidth]{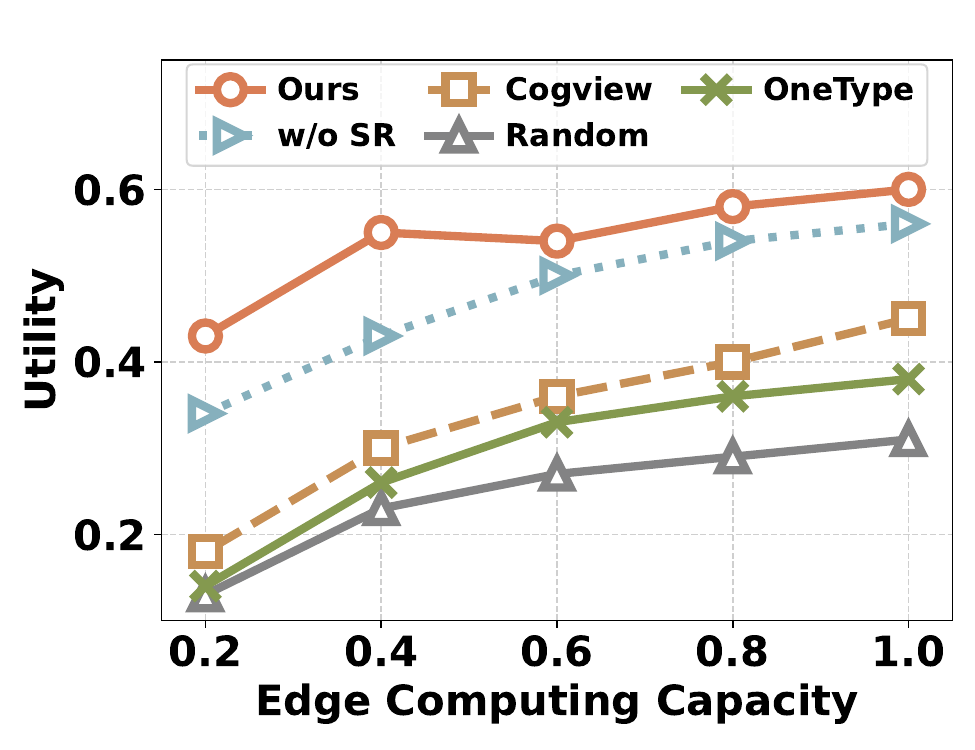}}
     \subfigure[The patch allocation ratio]{
            \includegraphics[width=0.49\linewidth]{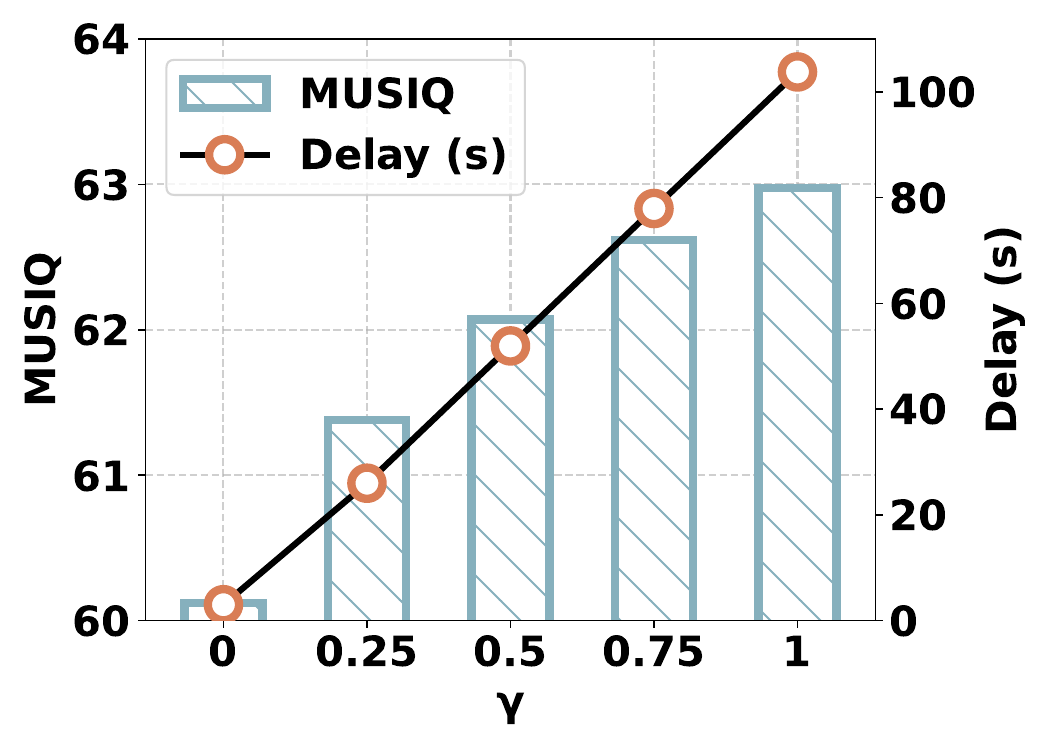}}

        \caption{The impact of edge available computing capacity and patch allocation ratio on the utility.}
    \label{fig:result2}
    \vspace{-0.3cm}
\end{figure}

We then investigate the effect of the patch allocation ratio $\gamma$ in the hybrid SR stage. A ratio of 0.25 indicates that the top-variance quarter of blocks is refined by the diffusion-based SR to better recover fine details, while the remaining blocks are enhanced by the learning-based SR for faster upscaling. As shown in Fig. \ref{fig:result2}(b), image quality across different $\gamma$ is evaluated using MUSIQ\cite{ke2021musiq}. Unlike CLIPScore which measures text-image alignment, MUSIQ reflects the fidelity of visual details based on a multi-scale Transformer that predicts perceptual quality from a single image and correlates well with human opinion scores. The result shows that overly large ratios incur excessive SR latency, whereas overly small ratios undercut detail recovery. Balancing these factors, we adopt 0.25 as the block-allocation ratio in this work.

\section{Conclusion}

In this paper, we have proposed an end-edge collaborative framework with adaptive configuration selection that minimizes service latency for high-resolution T2I generation tasks in resource-constrained edge environments. Also, We have introduced a region-aware hybrid SR policy that decouples the enhancement process. Specifically, low-resolution images generated by the edge T2I model are partitioned by spatial variance, with foreground regions enhanced by a powerful diffusion-based SR model on the edge server to ensure high image quality, while background regions are efficiently upscaled by a lightweight SR model on the user device. Experimental results have demonstrated that our framework has reduced service latency by 33\% while maintaining image quality comparable to baselines.

\vspace{12pt}


\bibliographystyle{IEEEtran}
\bibliography{ref}

\end{document}